# AN ADAPTIVE STRATEGY FOR THE CLASSIFICATION OF G-PROTEIN COUPLED RECEPTORS

S. Mohamed, D. Rubin, and T. Marwala*

*School of Electrical and Information Engineering, University of the Witwatersrand, Johannesburg. Private Bag 3, Wits, 2050, South Africa.

**Abstract:** One of the major problems in computational biology is the inability of existing classification models to incorporate expanding and new domain knowledge. This problem of static classification models is addressed in this paper by the introduction of incremental learning for problems in bioinformatics. Many machine learning tools have been applied to this problem using static machine learning structures such as neural networks or support vector machines that are unable to accommodate new information into their existing models. We utilize the fuzzy ARTMAP as an alternate machine learning system that has the ability of incrementally learning new data as it becomes available. The fuzzy ARTMAP is found to be comparable to many of the widespread machine learning systems. The use of an evolutionary strategy in the selection and combination of individual classifiers into an ensemble system, coupled with the incremental learning ability of the fuzzy ARTMAP is proven to be suitable as a pattern classifier. The algorithm presented is tested using data from the G-Coupled Protein Receptors Database and shows good accuracy of 83%. The system presented is also generally applicable, and can be used in problems in genomics and proteomics.

## 1. INTRODUCTION

Biosequence analysis has received increased attention in recent years since the completion of the human genome project. As a sub-field, protein sequence analysis has also become important due to its application in drug discovery programs [1] and in the analysis of prion diseases. The benefit of a computational analysis of biological systems is most clear when analysing the process of drug design. The development of new drugs often takes up to 15 years and costing up to $700 million per drug under investigation [1]. This drug design consists of two phases: a discovery phase and testing phase [2]. It is in this drug discovery phase that computational tools have had the most impact. In pharmaceutical drug discovery programs it is often useful to classify the sequences of proteins into a number of known families. In a mathematical notation, if it is known that a sequence $S$ is obtained for some disease $\mathcal{X}$, and that $S$ belongs to family $\mathcal{F}$, treatment for the disease is initially determined using a combination of drugs that are known to apply to $\mathcal{F}$ [3].

Consider the example of the HIV protease, a protein produced by the human immunodeficiency virus. The target identification stage involves the discovery of this HIV protease and the identification of this protein as a disease causing agent. The objective of drug design is to design a molecule that will bind to and inhibit the drug target. A great deal of time and money can be saved if the effect of molecules can be determined before these molecules are actually synthesised in a laboratory. Bioinformatics tools are used to predict the structures and hence the functions of the molecules under design and to determine if they will have any effect on the drug target.

The G-Protein Coupled Receptors (GPCRs) are the most important superfamily of proteins found in the human body. Many classification systems have been developed over the years based on machine learning to classify sequences as belonging to one of the GPCR families, and have shown great success in this task. These classification systems produce static classifiers which cannot accommodate any new sequences that may be discovered.

This paper introduces the use of a classification system based upon an evolutionary strategy, incremental learning and the Fuzzy ARTMAP to realise a protein classification system for the GPCR protein superfamily that allows all-vs.-all comparison of these proteins. Being an incremental system, the classifier is dynamic and has the ability to incorporate new information into the classification model.

## 2. IMPORTANCE OF GPCRS

The G-Protein Coupled Receptors (GPCRs) are a superfamily of proteins and forms the largest superfamily found in the human body. The GPCRDB is a database dedicated to the storage and annotation of G-Coupled proteins and at present consists of 16764 entries [4]. GPCRs play important roles in cellular signalling networks in processes such as neurotransmission, cellular metabolism, secretion, cellular differentiation and growth and inflammatory and immune responses [5]. Because of these properties, the GPCRs are the targets of approximately 60% – 70% of drugs in development today [6], 50% of current drugs on the market and approximately 20% of the top 50 best selling drugs target GPCRs. This results in greater than US$23.5 billion in pharmaceutical sales revenue from drugs which target this superfamily [6]. GPCRs are associated with almost every major therapeutic category or disease class, including pain, asthma, inflammation, obesity, cancer, as well as cardiovascular, metabolic, gastrointestinal and

CNS diseases [7]. This obvious importance of the GPCRs is the reason they are used in this research.

The key features of the GPCRs are that they share no overall sequence homology and have only one structural feature in common [5]. The GPCR superfamily consists of five major families and several putative families, of which each family is further divided into level I and then into level II subfamilies. The extreme divergence among GPCR sequences is the primary reason for the difficulty of classifying these sequences [1], and another important reason as to why they are used in this research.

In this research eight GPCR families are considered from the number of families available in the GPCRDB. The GPCR sequences are stored in the EMBL format, which consists of a number of labelled fields considering aspects of a sequence such as identifiers in a number of databases, the date of discovery and relevant publications dealing with the protein sequence. The database itself is updated every three to four months.

The distribution of the sequence lengths in the data that is used is an important factor to consider. Figures 1 shows a histogram of the sequence length distribution for the data that is used and shows that the data has a unimodal distribution, with most sequences having a length of about 350 amino acids for the GPCR data. The distribution also shows that the data does include sequences of lengths both longer and shorter than that indicated at the mode. We can use this as an indication that the data used is sufficiently representative of the protein data in general and that results from experiments that are conducted can be used to show that the algorithms are not highly dependant on sequence lengths for classification.

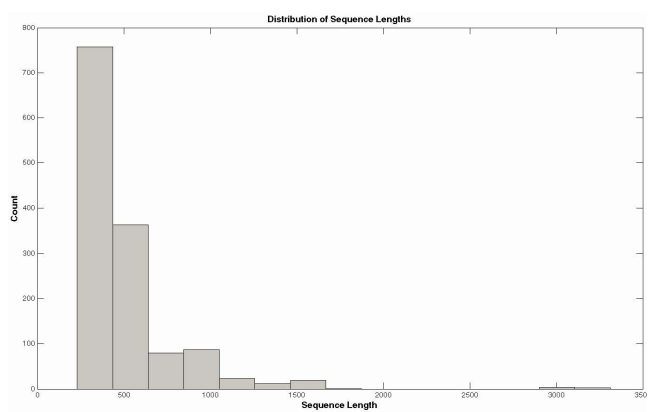

**Figure 1: Sequence length distribution for GPCR data**

### 3. SEQUENCE ALIGNMENT TECHNIQUES FOR CLASSIFICATION

Sequence alignment is the procedure of comparing two (pair-wise alignment) or more (multiple alignment) DNA or protein sequences by searching for a match between characters or groups of characters in each sequence [8]. The degree of similarity is described by a fractional value and there exits three categories of computational methods to perform these alignments.

The simple or **pairwise alignments** determine similarity by aligning a query sequence with every other sequence in a sequence database using an amino acid similarity matrix. Smith–Waterman [9] and Needleman–Wunsch algorithms [10] are dynamic programming techniques that find optimal local and global alignments respectively. Once an optimal alignment is determined, a scoring matrix is used which allows us to determine the degree of similarity between the aligned sequences. While the algorithms are efficient in determining the optimal alignment between two sequences, it becomes computationally infeasible for use in a database-wide search. This problem though has been overcome by a number of heuristic database search techniques such as BLAST [11] and FASTA [12], which have become more prevalent and efficient for database-wide searches.

The **multiple alignments** search against a database of known sequences by first aligning a set of sequences from the same protein superfamily, family or subfamily and creating a consensus sequence to represent the particular group. The query sequence is then compared against each of the consensus sequences using a pairwise alignment. The query sequence is classified as belonging to the group with which it has the highest similarity score [1]. Some popular techniques for performing multiple sequence alignments are Position Specific Scoring Matrices (PSSM) [13] and ClustalW [14]. The third category uses **profile Hidden Markov Models** (HMMs) as an alternative to the consensus sequences, but is otherwise identical to the multiple alignment technique. The focus of this research is not on alignment based techniques and thus they are not described in detail here. The alignment based techniques are described in detail in [2, 8, 15, 16].

### 4. PROBLEMS WITH ALIGNMENT BASED TECHNIQUES

Many shortcomings have been identified with respect to the effectiveness of sequence alignments, which is the reason why these techniques are not considered here. The principle argument against sequence alignment is the assumption that the order of homologous segments is conserved [17]. This assumption contradicts accepted understanding that evolution causes genetic recombination and reshuffling of nucleotides and amino acids [18]. The other argument lies in the lack of computational efficiency of the approaches.

This has led to the development of so called "alignment–free" techniques. These techniques rely mainly on machine learning approaches [19] and the application of Information theory, Kolmogorov complexity and Chaos theory [17]. Popular machine learning tools that have been applied to problems in protein classification include the Multi–layer Perceptron neural networks [20, 21], Support Vector Machines [22, 23], k-Nearest Neighbour Classifiers [24] and Naive Bayes Classifiers [1], among others.

A pattern recognition approach is adopted in this research to classify protein primary structures into a number of primary and putative families. The pattern recognition approach allows the time complexity to be limited to the initial training procedure and does not make any assumptions as to the order of homologous segments of a protein.

## 5. CURRENT CLASSIFICATION TOOLS IN USE

The feature based approach to protein sequence classification makes possible the use of a wide range of classification tools. Most protein databases supply Hidden Markov Models (HMM) for each of the families in the database, and the HMM's can be used to determine which family an unknown sequence belongs to. More recently, the use of Multi–Layer Perceptron (MLP) Neural Networks has been introduced to the problem of classification. Neural networks have been applied by authors such as Dubchak [25], Nagarajan *et al* [26] and Weinert and Lopes [21]. Each has shown success in the areas of domain detection or protein folding prediction. Other types of classifiers have also been used. Zhao *et al* [27] have made use of the Support Vector Machines while Radial Basis Function (RBF) Neural Networks and *k*-Nearest Neighbour (*k*-NN) classifiers have also been used [24].

### 5.1 Fuzzy ARTMAPs for Classification

This paper introduces the Fuzzy ARTMAP as a classifier for the protein classification task. The fuzzy ARTMAP is based on adaptive resonance theory and was introduced by Carpenter *et al* [28]. This learning system is built upon two fuzzy ART modules and employs calculus based fuzzy operations in the learning procedure. A diagram showing the structure of a fuzzy ARTMAP system is shown in figure 2.

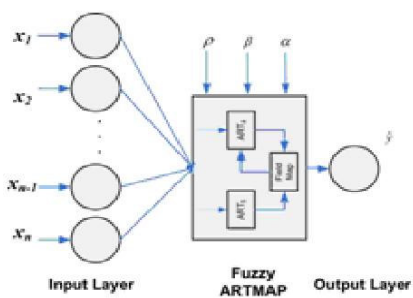

**Figure 2: Representation of the Fuzzy ARTMAP Architecture**

The fuzzy ARTMAP divides the input feature space into a number of hyperboxes in the *n*-dimensional space. It contains a map field which maps the individual hyperboxes to the output classes of the classification system. As a result, the fuzzy ARTMAP is able to model complex input spaces well. It requires two variables, where the vigilance parameter $\rho$, represents the tradeoff between classification accuracy and incremental learning ability. The learning rate $\beta$, is a factor by which the hyperboxes are adjusted with each training pattern during the training phase. In this system, $\beta = 1$, which is known as fast learning. Further details on the fuzzy ARTMAP and its training can be found in [28].

### 5.2 Overview of the Genetic Algorithm

Genetic algorithms (GA) find approximate solutions to problems by applying the principles of evolutionary biology, such as crossover, mutation, reproduction and natural selection [29]. The GA search process consists of the following steps: 1) generation of a population (pool) of candidate solutions, $\mathbf{s} = \{s_1, s_2, \ldots, s_p\}$, where *p* is the size of the population. 2) Evaluation of the fitness for each chromosome $s_i$ in the gene pool. Chromosomes with the lowest fitness are discarded and make way for a new set of chromosomes. Replacement sets of chromosomes are created by the genetic operations of crossover and mutation on the most fit individuals. 3) Steps 1 and 2 are repeated for a given number of generations until a specified fitness level is attained or a maximum number of generations are exceeded [30].

The genetic algorithms represent input data from the problem by an encoding such as binary or floating point and use the genetic operations to iteratively evaluate solutions from the population of potential solutions to determine the global optimum [30]. The GA evaluates candidate solutions through a fitness function and by maximising this fitness function, determines the global maximum. The fitness function contains information from the problem space and is the mechanism by which properties of the problem space is transferred to the GA, which is independent of the problem. The genetic operations are important since they add an element of randomness to the search process, allowing a wider range of the solution space to be explored.

## 6. PRIOR WORK

The problem of incremental learning has not been considered before as it is presented here. Vijaya et al [31] consider the incremental clustering of protein sequences, but that is a different problem from that considered here. The fuzzy ARTMAP has been chosen as the incremental classifier and as mentioned, has been shown to be an effective incremental classifier [28]. The Support Vector Machine (SVM) is widely used in protein classification and it would appear that the use of an incremental SVM would be more suitable. While some algorithms for incremental SVM [32] exist, the problem with many of these systems is that they cater to the binary-classification problem only and are not applicable to multi–class classification problems, which is the case for the classification of proteins into families. Other incremental classification systems also exist, such as incremental common-sense models and incremental fuzzy decision trees. Of these incremental classification systems, the fuzzy ARTMAP is the most established and well known and is thus used.

## 7. SYSTEM OVERVIEW

A schematic representation of the system is shown in figure 3. Input sequences are extracted from a protein database and then converted into a numerical feature vector. We then create a population of classifiers to introduce classification diversity, with the selection of suitably diverse classifiers from this population using the Genetic Algorithm coupled with kappa analysis. An ensemble of classifiers is used as a means of introducing modularity in the learning system. This system is implemented using the fuzzy ARTMAP (FAM) and a series of experiments are conducted to evaluate the performance of this system. Pseudocode for the creation and operation of the system is shown in algorithm listing 7. The ability of the FAM as an alternative classifier to many of the other more popular classifiers is demonstrated by comparing the classification ability of these systems using the GPCR data set. The incremental learning system described by algorithm listing 7 is then tested using the GPCR data and shown to be able to learn new data as well as maintain existing data.

**Algorithm 7.1:** FUZZY ENSEMBLE($D$)

**Training Phase**
- **comment:** Create population $j$ of FAM classifiers each trained with a different permutation of the input data $\mathbf{X}_1$
- Each classifier is a hypothesis $h_t : \mathbf{X}_1 \mapsto \mathbf{Y}_1$
- $\epsilon = \frac{1}{N} \sum n|_{h_t(x_i) \neq y_i}$
- **comment:** Sort classifiers based on incr. error on validation set.
- SORT($\epsilon$)
- **comment:** Select lowest error classifier as elite classifier $h_{elite}$
- **comment:** Calculate the agreement $\kappa$, of the 15 best classifiers (based on error) with respect to the elite classifier
- $\kappa = \frac{N \sum_{i=1}^{N} x_{ii} - \sum_{i=1}^{N} x_{i+} \cdot x_{+i}}{N^2 - \sum_{i=1}^{N} x_{i+} \cdot x_{+i}}$
- **comment:** Genetic Algorithm selection of $p$ classifiers based on a trade-off between error $\epsilon$ and agreement $\kappa$
- $GA_{fitness}(\kappa, \epsilon) = \lambda \sum_{i=1}^{p} \kappa_i + \sum_{i=1}^{p} \epsilon_i$
- Create ensemble classifier using the elite classifier $h_{elite}$ and the $p$ selected classifiers $h_t, t = 1, \ldots, p$
- **comment:** Fusion of individual predictions using majority voting.

**Operation Phase**
- If predicting sequence family, convert to feature representation and classify using the Fuzzy ARTMAP based system created during this previous training phase
- **comment:** If incrementing system knowledge, increment each classifiers in the Fuzzy ARTMAP base system independently, using the training data for new sequences
- $h_t^{incr} = \mathcal{T}(h_t, \mathbf{X}_k \mapsto \mathbf{Y}_k)$,
- where the transformation $\mathcal{T}$ is the incremental training process and $k$ is the dataset to be added to the system

## 8. PROTEIN VECTORISATION

The data obtained from the GPCRDB is in the form of amino acid sequences. In order for these sequences to be used in classification systems, they must be converted into a numerical form. Before this conversion though, preprocessing in the form of outlier removal must be completed. Outlier removal consists of removing sequences which have characters which are not part of the standard 20-letter amino acid alphabet — the letters are B and Z and have ambiguous meanings. Once this process is complete, these protein sequences must be transformed into numerical features. Two types of features have been identified in the literature, these being global and local features. Huang et al [33] provide a good description of the difference between global and local features and this distinction is used in this work.

### 8.1 Global Feature Generation

Global features represent the nature of the entire protein sequence. These features must capture the global similarity between related sequences allowing for comparison. Consider the amino-acid composition of the sequence. The composition is simply the presence frequency of each of the 20-possible amino acids in the given sequence. Thus the composition is calculated by [27]:

$$\nu_i = \frac{s_i}{\sum_{j=1}^{20} s_j}, \text{ for } i = 1, 2, \ldots, 20.$$

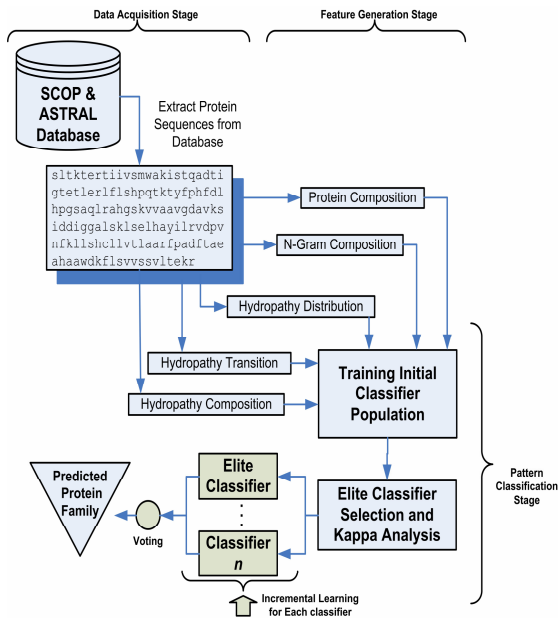

**Figure 3: Overview of System Architecture**

where $\nu_i$ is the value for the $i$th feature and $s_i$ is the number of times the $i$th amino acid appears in the sequence. This results in 20 features: a frequency of appearance for each of the possible amino acids. If a particular amino acid does not appear at all in the sequence, the corresponding feature value is zero.

A second set of features based on the hydropathy of amino acids in a given protein sequence is also calculated. Amino acids are either hydrophobic, hydrophilic (polar) or neutral. We use the Chothia and Finkelstein [25] hydropathy classification. We calculate three descriptors, the hydropathy composition (**C**), the hydropathy distribution (**D**) and the Hydropathy transmission (**T**) for the sequences as described by Dubchak [25].

The composition **C**, is calculated similarly to the amino acid composition described previously. In this case we calculate the presence frequency of hydrophobic, hydrophilic and neutral amino acids in the sequence. This results in three features being generated. The transmission **T**, is defined by three values. The first is the number of times a polar molecule I followed by a neutral molecule or vice versa. Similarly the other two are the number of times a neutral molecule is followed by a hydrophobic molecule or vice versa and the number of times the polar molecule is followed by a hydrophobic molecule or vice versa.

The distribution **D**, looks at intervals of 25%, 50%, 75% and 100% along the sequence length. For each interval the presence frequency of hydrophobic, hydrophilic and neutral molecules for each percentage interval is calculated. This results in 12 features, 4 features for each of the three hydropathy groups. A more detailed description of these features can be found in Dubchak [25]. In total 38 features (20+3+3+12) are generated based on global sequence descriptors.

*8.2 Local Feature Generation*

The local features capture local interactions between amino acids and groups of amino acids in a protein sequence. The *n*-gram method is well established as a good descriptor of local similarities in a sequence and has been used by many authors such as Cheng et al [1], Tomovic et al [23] and Zhao et al [19]. Essentially the n-gram method considers the presence frequency of consecutive n-letter combinations in the protein sequence, for integer *n*. For example, consider the short sequence `SLTKTERTIIVSM`, the 2-grams of this sequence are: `SL, LT, TK, KT,` etc. Given a sequence, features are generated by calculating the presence frequency of all possible n-grams for the amino acid alphabet. Two letter combinations are known as digrams or bigrams. While higher n-grams such as 3-grams and 4-grams have been considered in the available literature, only digrams are considered in this work since it has been proven by numerous authors [1, 19] to work well in protein classification systems.

A total of 438 features have been generated and as a final post-processing step, the features undergo min-max normalisation. The normalisation is a requirement for using the FAM, since the FAMs complement coding scheme assumes normalised data.

9. INCREMENTAL ALGORITHM AND DIVERSITY

The creation of the committee–based system is based on a novel approach, implementing an evolutionary strategy which was summarised in the algorithm listing. We first train an initial population of *j* classifiers, each classifier having been trained with a different permutation of the input training data. This permutation is needed in order to add diversity to the classifiers being created. As mentioned, the fact that the fuzzy ARTMAP learns in an instance–based fashion, makes the order in which the training patterns are received an important factor [34]. In the experiments performed, the initial population consists of 30 classifiers.

The classification error $\epsilon$, of each of these classifiers is then evaluated against a validation data set. The classifiers are then ranked in terms of increasing error. The lowest error classifier from this population is the elite classifier and is the classifier that automatically becomes a member of the ensemble system. The inclusion of this elite classifier ensures that at least one high accuracy classifier is selected for the committee. The next step is to select the remaining *n* classifiers. In this application we select a further 4 classifiers. The selection of the other members of the committee is important and requires a number of factors to be considered:

- We do not wish to select classifiers that perform exactly as the elite classifier, since this gives no diversity to the predictions that are generated, and thus there is no room for improvement.
- We do not wish to select low accuracy classifiers that will confuse the prediction obtained and thus result in predictions that are more erroneous than a single classifier.

It would appear that these two conditions oppose each other, since high accuracy classifiers would tend to agree on the same predictions, against what we require for point 1. A trade-off between the classifier accuracy and the level of agreement between classifiers is then ideally what is required. This introduces the need for a formal definition of agreement between classifiers.

We use the definition of agreement considered by Petrakos et al [35], and the mathematical description that follows is generally known as kappa analysis. We define the agreement between any two classifiers $\kappa$ based on the error matrix of the two classifiers [36]. The error matrix shows the number, and for which classes the two classifiers agree on a prediction. Table 1 shows the format for an error matrix between two classifiers.

Table 1: Data Format for Error Matrix Between Classifiers

| Classifier 1 | Classifier 2 | | | | |
|---|---|---|---|---|---|
| | $C_1$ | $C_2$ | ... | $C_Q$ | Totals |
| $C_1$ | $x_{11}$ | $x_{12}$ | ... | $x_{1Q}$ | $x_{1+}$ |
| $C_2$ | $x_{21}$ | $x_{22}$ | ... | $x_{2Q}$ | $x_{2+}$ |
| ⋮ | ⋮ | ⋮ | ⋱ | ⋮ | ⋮ |
| $C_Q$ | $x_{Q1}$ | $x_{Q2}$ | ... | $x_{QQ}$ | $x_{Q+}$ |
| Totals | $x_{+1}$ | $x_{+2}$ | ... | $x_{+Q}$ | |

In the above table, $Q$ is the number of classes in the data. $x_{11}$ in the table is the number of test patterns that both classifier 1 and 2 agreed belonged to class $C_1$. $x_{21}$ is the number of test patterns that classifier 1 predicted belonging to class $C_2$, but that classifier 2 predicted belonged to class $C_1$. Similarly, the entire error matrix can be generated using the prediction made by any two classifiers. We determine the error matrices for 15 of the best classifiers in terms of predictions with respect to the elite classifier. The agreement is calculated using the following set of equations, where N is the number of training patterns used in generating the error matrix [36].

$$\theta_1 = \sum_{i=1}^{N} x_{ii}$$
$$\theta_2 = \sum_{i=1}^{N} x_{i+} \cdot x_{+i}$$
$$\kappa = \frac{N\theta_1 - \theta_2}{N^2 - \theta_2}$$

The selection of classifiers from this population, which must essentially minimise both the error of the individual classifiers and the agreement of the classifiers with the elite classifier, is an optimisation problem. We have chosen to implement a Genetic Algorithm as the optimisation tool for this system. The Genetic Algorithm (GA) is a stochastic optimisation tool that borrows concepts from evolutionary biology such as selection, crossover and mutation [37]. The GA minimises a cost function that is defined for a particular problem by stochastically exploring the space of available solutions. The GA implemented for the selection of classifiers is designed to select 4 classifiers and minimises both the agreement and the error of the selected combination of classifiers.

The GA will select 4 classifiers, resulting in two vectors:

$$\epsilon_{\mathbf{GA}} = \{\epsilon_1; \epsilon_2; \epsilon_3; \epsilon_4\}$$
$$\kappa_{\mathbf{GA}} = \{\kappa_1; \kappa_2; \kappa_3; \kappa_4\}$$

We use a linear combination of these two matrices to define the cost value of a particular selection of classifiers. It is this cost that the GA will attempt to minimise. The cost function is defined by equation 5. $\lambda$ is introduced as a scalar constant to allow the relative importance of the agreement in the system to be adjusted. In this study $\lambda = 1$, which gives equal importance to both the error and the agreement.

$$f(\epsilon, \kappa) = \lambda \sum_{i=1}^{4} \kappa_i + \sum_{i=1}^{4} \epsilon_i$$

The GA selects the 4 best classifiers that minimises the cost function of equation 5. The Genetic Algorithm was designed to produce 50 generations of solutions with each generation being a population 30 possible solutions. The crossover rate was set to a high value of 0.8 and a mutation rate of 0.4, and was empirically determined to be the best values for the experiment. The crossover functions are modified from the standard crossover functions in this case, to ensure that unique classifiers are selected during each generation, that is, preventing the same classifier from being selected twice in a particular generation.

These selected classifiers are then used in parallel, with each of the five classifiers in the system producing an independent set of predictions. These predictions must then be fused together to form the final decision. A number of decision fusion techniques exist. Some of these include the majority and weighted majority voting, trained combiner fusion, median, min and max combiner rules [38]. We adopt the majority voting decision fusion scheme, which simply considers each of the predictions produced by the five classifiers as a vote, with the final prediction for any given pattern given by the prediction that receives the largest number of votes.

### 9.1. Incremental Learning of Protein Data

The ensemble system is not a useful system if it is not able to accommodate newly discovered sequences that are produced daily. The ability of a classifier to allow this type of knowledge update was also defined as incremental learning. The fuzzy ARTMAP through its instance–based learning is able to incrementally learn new data. This incremental learning can consider two types of data:

1. It is possible to add new sequence information for families which the classifier has already been trained with.
2. Data of completely new classes can be added to the system, increasing the knowledge that the system has of the general protein domain.

The base system will in general be trained with data of a number of classes. Once new data becomes available, incremental learning of the system is based on incrementally training each of the 5 FAM classifiers in the system with the new data. The system can now be tested with data from all classes it has been trained with, including classes which have been incrementally added to the system.

## 10. SYSTEM TESTING AND EXPERIMENTAL RESULTS

### 10.1. Testing Using GPCR Data

The GPCR data is also divided into 6 separate databases $\mathcal{D}_1, \ldots, \mathcal{D}_6$, with a validation set for database $\mathcal{D}_1$. In this case, the datasets have data of all 8 classes which are available. This specific partitioning is used to demonstrate data incremental learning, where new data of classes which the system has already been trained with is added to the system. This case is more appropriate for use with GPCR data where the families are established. The separation of data into these databases is shown in table 2.

Table 2: Separation of data into individual databases for testing using GPCR data. $\mathcal{D}_v$ and $\mathcal{D}_t$ are the validation and testing datasets respectively.

| Family | $\mathcal{D}_1$ | $\mathcal{D}_v$ | $\mathcal{D}_2$ | $\mathcal{D}_3$ | $\mathcal{D}_4$ | $\mathcal{D}_5$ | $\mathcal{D}_6$ | $\mathcal{D}_t$ |
|---|---|---|---|---|---|---|---|---|
| Type 1 | 32 | 10 | 43 | 43 | 43 | 43 | 43 | 43 |
| Type 2 | 23 | 8 | 30 | 30 | 30 | 30 | 30 | 30 |
| Type 3 | 16 | 6 | 22 | 22 | 22 | 22 | 22 | 22 |
| Type 4 | 6 | 2 | 9 | 9 | 8 | 8 | 8 | 8 |
| Fz/Smo | 12 | 4 | 16 | 15 | 16 | 16 | 16 | 16 |
| MLO | 3 | 1 | 4 | 5 | 5 | 5 | 5 | 4 |
| Class H | 32 | 11 | 43 | 43 | 43 | 43 | 43 | 43 |
| Pheromone 2 | 20 | 6 | 26 | 26 | 26 | 26 | 27 | 27 |

### 10.2. Comparative Performance

We compare the Fuzzy ARTMAP with other more common machine learning tools such as the Support Vector (SVM) Machines and Multi-layer perceptron (MLP). These have been chosen since they have found widespread use in the literature [1, 3, 19]. Table 3 shows the performance of the classifiers that were considered in the experiment. The parameters that are used for each of the classifiers is included in the table. The classifiers are trained with all the training data combined into a single training set and tested on the test set $\mathcal{D}_t$, using the features that were described in section 5. The table shows that the FAM has comparable accuracy when compared to many other classification systems.

Table 3: Comparative performance of FAM versus other classifiers on the GPCR dataset.

| Classifier | Error (%) |
|---|---|
| Generalised Linear Model | 25.91 |
| Multi–layer Perceptron, $n_{hid} = 15, cyc = 200$ | 15.03 |
| Fuzzy ARTMAP $\rho = 0.75$ | 11.90 |
| SVM - RBF $\gamma = 2.3$ | 17.10 |
| SVM-Polynomial 2.23 degree | 10.36 |

### 10.3. Base Classifier Training and Incremental Performance

The base classification system was trained using database $\mathcal{D}_1$. Table 4 shows the error of the first 15 classifiers of the population and agreement with the elite classifier. The error is the error of the system on the validation data set. The GA for this data set selected classifiers `2,3, 4, and 12` to form the final ensemble system. Again, the system consisting of the elite classifier and the four classifiers selected by the GA are incrementally trained using databases $\mathcal{D}_2, \ldots, \mathcal{D}_6$ with the ensemble being tested after each increment with the testing database $\mathcal{D}_t$. The performance of the system is shown in table 5.

This data shows that the system is extremely capable of remembering data that has been trained upon, as shown by the many 0% which appear in the table for the training databases. The many zeros are not an indication of overtraining. The FAM is trained so that it learns all its training data with a 0% error. What the results show is that after it has learnt its initial training data, the memory is not degraded by the addition of additional data. The system also shows that the performance does increase as more data of each of the classes is added to the system.

Table 4: Error and Agreement values for 15 classifiers of the population

| Classifier | Val Error $\epsilon$ (%) | Agreement $\kappa$ |
|---|---|---|
| 1 | 27.0833 | Elite |
| 2 | 29.1667 | 0.8940 |
| 3 | 29.1667 | 0.9730 |
| 4 | 29.1667 | 0.8438 |
| 5 | 31.2500 | 0.8929 |
| 6 | 31.2500 | 0.8929 |
| 7 | 31.2500 | 0.8929 |
| 8 | 31.2500 | 0.8455 |
| 9 | 31.2500 | 0.8683 |
| 10 | 31.2500 | 0.8929 |
| 11 | 31.2500 | 0.8929 |
| 12 | 31.2500 | 0.8929 |
| 13 | 31.2500 | 0.8929 |
| 14 | 31.2500 | 0.8430 |
| 15 | 33.3333 | 0.8430 |

Table 5: Training and generalisation performance of system on GPCR data

| Set | Train 1 | Train 2 | Train 3 | Train 4 | Train 5 | Train 6 |
|---|---|---|---|---|---|---|
| $\mathcal{D}_1$ | 0 | 0 | 0 | 0 | 0 | 0 |
| $\mathcal{D}_2$ | — | 0 | 0 | 0 | 0 | 0 |
| $\mathcal{D}_3$ | — | — | 0 | 0 | 0 | 0 |
| $\mathcal{D}_4$ | — | — | — | 0 | 0 | 0 |
| $\mathcal{D}_5$ | — | — | — | — | 0 | 0 |
| $\mathcal{D}_6$ | — | — | — | — | — | 0 |
| $\mathcal{D}_v$ | 25.00 | 22.92 | 22.92 | 27.08 | 25.00 | 27.08 |
| $\mathcal{D}_t$ | 22.79 | 18.65 | 19.17 | 19.69 | 18.65 | 16.58 |

## 11. ANALYSIS OF RESULTS

We have described the tools and techniques that are currently used in the classification of protein primary structures into families and the introduction of two algorithms for incremental learning of this protein data. There has been a great deal of work in the classification of these proteins using a wide range of computational intelligence techniques ranging from the *k*-Nearest Neighbours classifiers and Naive Bayes classifiers to more complex tools such as the Multi–layer perceptron and the Support Vector Machines. While these systems have allowed a wider set of evolutionary mechanisms involving proteins to be included in the design of classification systems, such as invariance to the order of amino acid motifs in a sequence, they remain static structures which cannot incorporate newly discovered proteins into their models.

With this in mind, *Incremental Learning* was proposed as a machine learning approach to the classification of proteins. The system presented is based on an evolutionary strategy and the fuzzy ARTMAP classifier. The results presented indicate that the fuzzy ARTMAP is a suitable machine learning tool for the classification of protein sequences into structural families, which is comparable to many of the more established tools. An analysis of the sequences also showed that the system is able to classify proteins of varying lengths, and thus the length of the protein sequences used is not important.

The results presented indicate that the fuzzy ARTMAP is a suitable machine learning tool for the classification of protein sequences into structural families, which is comparable to many of the more established tools. The accuracy of the classification could be improved if some form of dimensionality reduction or feature selection is applied. These techniques have been applied by many authors using numerous techniques. Principal Component Analysis has been used as a technique of dimensionality reduction by Zhao et al [27] and Cheng et al [1] uses the chi-squared test as a means of feature selection. Feature selection can also be applied using various sub-optimal feature selection techniques such as the floating forward selection search using the $J_3$ measure as the distance function [39] or the Genetic Algorithm can be used as demonstrated by Mohamed *et al* [40].

For the fuzzy ARTMAP based system, the agreement κ was used to measure diversity of the system. The use of the correlation coefficient or the use of a *disagreement* [36] should also be explored, to determine if these alternate measures gives some degree of refinement in the selection of the classifiers. The genetic algorithm is also important in the committee. Due to the stochastic nature of the GA, it is possible that different GA optimisations produce a different selection of classifier members. This though is not as likely in the case of the data presented here, since many of the classifiers had the same agreement or error, resulting in the GA converging to the same selection choice. That said, the optimisation of the GA is efficient and runs very fast due to the fact that it uses pre-calculated results such as the error matrix and agreement values. It might seem that the contribution of the GA is not significant if the case of the testing using the GPCR data is considered. This might be the case for this data, but the algorithm is designed to be generally applicable, and thus this might not be the case for another set of data, which also need not necessarily be protein data.

## 12. CONCLUSION

Initial researchers into incremental learning such as Elman [41] claimed that incremental learning is always superior to batch learning. We choose to adopt a softer approach and rather emphasise that, while the batch trained approach may be suitable, the incremental approach save a great deal of time and allows previous classifier design effort to be maintained. Where the case exists that any new information that may be obtained will not significantly improve the classification ability of the system, then the batch training approach may be more suitable. Where this is not the case such as families whose sequences have low sequence similarity, then the incremental approach may be better and will be more desirable.

The algorithm presented is applicable in general to all classification problems and is not limited to the problem of structural family classification. The algorithm can be easily extended to secondary and tertiary structure prediction, functional annotations and the prediction of protein–protein interaction sites. Apart from systems in proteomics, genomic applications also exist, such as the classification of promoter, and splice sites. Each classification task benefits from the improvements which can be gained from using an ensemble system and incremental learning. These results show great promise for the future of computational biology, where newly discovered data needs to be accurately incorporated into existing models, allowing for highly agile discovery processes.